\pdfoutput=1

\documentclass[11pt]{article}

\usepackage{emnlp2023}

\usepackage{times}
\usepackage{latexsym}
\usepackage{amsmath} 
\usepackage{booktabs}
\usepackage{graphicx}
\usepackage[T1]{fontenc}

\usepackage[utf8]{inputenc}

\usepackage{microtype}

\usepackage{xspace}
\usepackage{comment}

\usepackage{inconsolata}

\title{MIL-BERT: Classification of Arbitrarily Large Text with Performance and Explanatory Guarantees}

\author{John Cadigan \\ SRI International \\
  \texttt{john.cadigan@sri.com} \\\And
  Dayne Freitag \\
  \texttt{dayne.freitag@sri.com} \\\And
  Eric Yeh \\ SRI International \\
  \texttt{eric.yeh@sri.com}}

\begin{document}
\maketitle
\begin{abstract}
Many text classification decisions are viable based on constituent excerpts alone. Taking inspiration from the field of multiple instance learning, we present an algorithm for training a neural network to classify text by selecting such excerpts. We show that our approach is also scalable with demonstrated learning against samples with nearly 1M tokens. We evaluate our methods on 7 datasets with emphasis on long-textual collections that far exceed the encoding limit of our base model. We present state-of-the-art results with this algorithm on 3 datasets: identification of political bias in news outlets, trigger warnings in long stories, and demographic characteristics of authors in tweet collections. Furthermore, the model trained on weakly-labeled collections of text (bags) generalizes to accurately classify constituent, smaller instances. Besides a new state-of-the-art for these problems, this approach is one of the few neural methods to excel in these datasets.
\end{abstract}

\newcommand{\candr}{C\&R\xspace}

\section{Introduction}

In the current state of the art, the most accurate approaches for textual classification and regression (\candr) are based on language models in the transformer family~\cite{vaswani2017attention}.  However, canonical transformers have fixed-length positional embeddings and computation and memory usage that grows quadratically with sequence length.  Many practical applications of \candr using popular foundation models make compromises such as truncation when processing long documents, implicitly assuming that sufficient information for accurate prediction can be found early in a document.  On the other hand, transformer adaptations for long documents such as Infinite Transformer~\cite{martins2022former} are relatively expensive to apply and process the entire text, opaquely and possibly needlessly.  We surmise that many problems of textual \candr can be successfully addressed by instead identifying one or a few passages that support a particular prediction---a phenomenon we call \emph{signal locality}.  For problems exhibiting signal locality, we can replace the wholesale processing of an overlarge text with a search for indicative excerpts of modest size.  This amounts to a \emph{multiple instance learning} (MIL) framing of the problem.

MIL is a paradigm of machine learning in which collections of \emph{instances} called \emph{bags}.  A bag is classified as positive as long as one of its instances is positive.  For example, MIL has been applied to the task of identifying cancerous image patches (instances) within larger, whole-slide images based on a slide-level (bag) label \cite{qu2024rethinking}. If an image contains cancer, then there is at least one image patch that contains cancer. If an image does not contain cancer, then no image patches contain cancer. An example application of MIL to text involves identifying civil unrest based on classifying collections of tweets for particular days and countries \cite{delucia2023multi}. 
Inspired by these approaches, we develop an algorithm for learning to classify arbitrarily long text with performance and explainability guarantees.

This paper makes the following contributions:
\begin{itemize}
    \item We describe an efficient approach for classifying long texts based on a MIL framing, involving a fixed set of ``selectors'' that identify the passages most relevant to a particular prediction.
    \item In experiments involving a range of previously proposed long-text classification problems, we demonstrate that this excerptive approach, MIL-BERT, is competitive with whole-document alternatives, in several cases surpassing previously reported accuracies.
    \item We describe the advantages of MIL-BERT for explainability, arising from the requirement that it identify key, easily reviewable passages that drive the classification decision.
\end{itemize}

\section{Method}

\subsection{Preliminaries}

In the canonical formulation of MIL~\cite{carbonneau2018multiple}, we perform binary classification of \emph{bags}, each bag $X=\{x_1,...x_n\}$ containing a set of instances $x_i$ with labels $y_i$.  A bag $X$ receives positive label $Y=1$ if any of its constituent instances $x_i$ has label $y_i=1$, in effect making MIL a sparse learning problem.  During training, only bag labels are given, and the objective is to learn an effective instance classifier.  Subsequent variants of MIL have prioritized bag labels~\cite{ilse2018attention} or presented noisy instance labels at training time ~\cite{qu2024rethinking}.  We explore problems with both visible (trigger warnings) and implicit instance labels (demographic profiling). 

\subsection{Algorithm}

We present a novel algorithm for textual \candr problems exhibiting signal locality, representing long documents as arbitrarily large bags of excerpts. In this study, we segment a document into excerpts $(x_1,x_2,...)$ with a rolling window and embed each excerpt with a RoBERTa model \cite{liu2019roberta}, using the embedding of the start symbol token (<s>) from the final layer. Our key innovation comes from the insight that, during training, intermediate activations of instances are kept in memory to later calculate the gradients even if they are not relevant to the gradients such as with a max pool shown in equation \ref{eq:double-pass}. 

\begin{equation}\label{eq:double-pass}
\small
\nabla max(f(x_1,...)) = \nabla max(f(argmax(f(x_1,...))))    
\end{equation}

\begin{figure}[h]
    \centering
    \includegraphics[ scale=0.15]{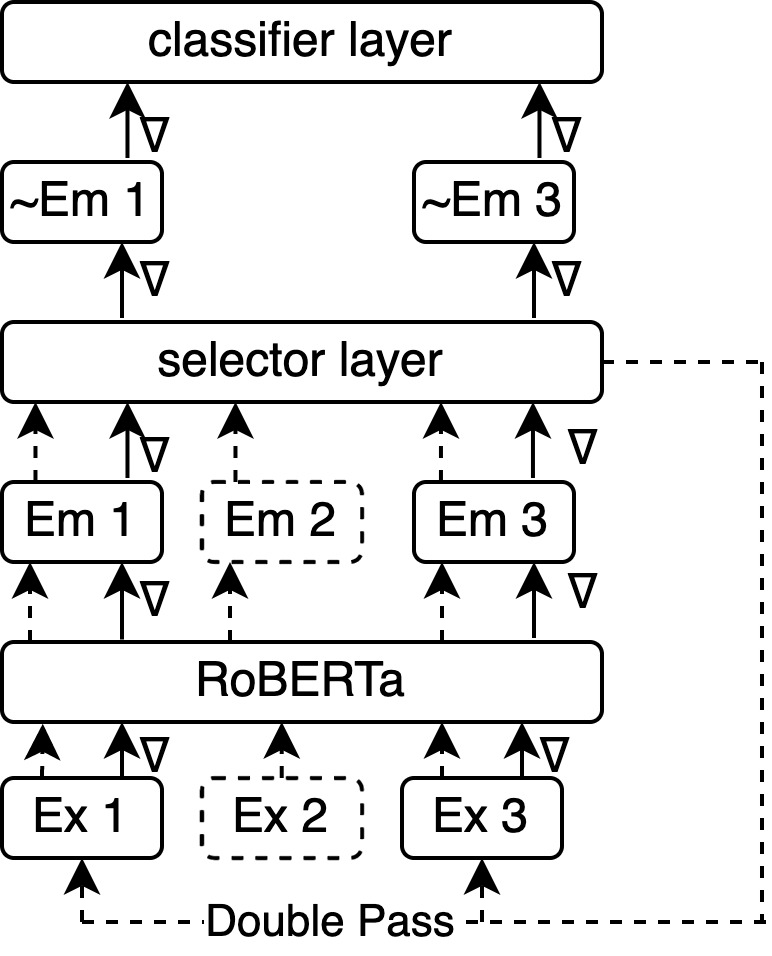}
    \caption{Documents are broken up into excerpts which are encoded with a BERT-style model. In the first pass (dotted line), a selector layer finds extreme points without gradients, saving memory. On a second run with gradients (dark lines), these are passed through the selector layer, a Gumbel-softmax operation, and onto a classifier layer. }
    \label{fig:double-pass}
\end{figure}

We achieve this mathematical substitution through what we refer to as the \emph{double-pass pooling trick}. In the first pass, without gradients, the model selects key excerpts with an argmax function we call a \textit{selector}. In the second pass, we run the model with gradients on the subset selected in the first pass. This allows the model to calculate the gradient for an arbitrary number of $n$ points with $k$ memory; the tradeoff is that we use $O(n+k)$ compute. We depict this process in Figure \ref{fig:double-pass}.

All selectors in this study are based on a linear layer of dimension $k$. We perform an affine transformation of all excerpts in the bag $(WX +b)$ and select the most extreme excerpts in each dimension, similar to a maxpool operation. There are multiple ways in which the excerpts thus selected can be combined to generate a prediction. Here, we consider two methods in this study. In the \textit{shared feature map} selector, parameters are shared between selection and classification, the output dimension of the affine transformation matches the number of classes and the top value for each dimension ultimately becomes the logit for that class. With variants selecting the top-k instances and aggregating scores \cite{delucia2023multi}, this classical MIL approach also learns an instance classifier from bag labels. In the \textit{feature tree} selector, each dimension of the affine transformation resembles a feature of a decision tree with a threshold of 0; if index 0 is greater than 0, then we next consider index 1; otherwise, index 2. This allows the use of $1 + 2^{k-1}$ selection decisions with only k instances selected, reducing memory usage. Geometrically, each selector index represents a hyperplane and the sign indicates which side of the hyperplane the bag is on, but this does not bear a direct relationship to classes as in the shared feature map selector. The \textit{feature tree} method loses instance-orientation to fit higher-order patterns. Other selectors are possible, but for double-pass pooling to work, selector methods must be \emph{idempotent}, i.e. application of selectors must yield the same result between the first and second passes. 



To enable a general range of classifier layers, we specify that selectors emit the embeddings of selected instances and pass them to the classifier. This requires an argmax operation which is not differentiable during training. Fortunately, the Gumbel-softmax trick allows a differentiable selection of individual instances \cite{jang2016categorical} by adding noise to the vector in order to create an approximately one-hot vector for the maximum value in the vector. This has been extended to perform top-k selection in a differentiable manner among other discrete operations \cite{paulus2020gradient}. This enables more powerful classification layers such as transformers or self-attention to be applied over instances as has been done in MIL research \cite{shao2021transmil}. For classifiers layers in this study, we use a simple linear layer but its dimensionality varies. The shared classifier layer shares all its parameters with its selector, using top-k excerpts to score each class with k equal to one. The feature tree classifier uses the values of selected indices as input into a linear layer ($1 + 2^{k-1} \times classes$) with the rest being zero. 

\section{Datasets}

We explore the capability of our algorithms on MIL with identifiable instance labels, MIL with implicit labels, and long-text documents. 

First, we investigate classic MIL applications where we train a classifier on bags of text and then evaluate its accuracy on \textit{both} bag and instance classification where labels are identifiable. The task is to identify trigger warnings in stories \cite{wiegmann2023trigger}, specifically the 2023 shared task \cite{wiegmann2023overview} \footnote{https://pan.webis.de/clef23/pan23-web/trigger-detection.html}. This dataset only has labels at the story (bag) level, but, in later work, they searched for passages (instances) within the stories that were likely to contain trigger words and labeled them with respect to 1 of 8 fine\-grained binary labels\cite{wiegmann2024if}; we evaluate instance accuracy for the 7 labels that match between the two datasets. The second MIL dataset comes from CLEF 2023 task 3 \footnote{https://gitlab.com/checkthat\_lab/clef2023-checkthat-lab/-/tree/main/task3}  which concerns identifying political bias in news sources (task B: bag) and individual articles (task A: instance) \cite{da2023overview}. Both of these benchmarks have standard training, validation and test partition data.

Second, we present results in demographic profiling of celebrity authors available from PAN2019@CLEF\footnote{https://pan.webis.de/clef19/pan19-web/celebrity-profiling.html} where gender, occupation, birthyear, and fame-level must be predicted based on tweet collections \cite{wiegmann2019overview}. Demographic profiling matches the use-case of MIL for learning over noisy data with implicit labels. The phrase "my husband" is indicative of gender, but it is not definitive, so excerpts in this domain defy explicit labeling. The dataset has a training and test partition, so we use 20\% of the training partition for validation with stratified sampling. For the birthyear task, we bucket years according to its task-specific F1-based scoring function that features wider buckets for older writers. Using MSE loss, we normalize the range by subtracting 1920 and dividing by 100. Anonymizing URL's improved results in this dataset.

As a final area of interest, we also consider the application of these methods to a long document classification benchmark \cite{park2022efficient} that contains results for previous algorithms: Hyperpartisan \cite{kiesel2019semeval}, 20Newsgroups, EURLEX-57K \cite{chalkidis-etal-2019-large}, and Book Text \cite{bamman2013new}. Train, validation and test partitions have been specified for each dataset \footnote{https://github.com/amazon-science/efficient-longdoc-classification}. Whereas the previous datasets have appeared as shared tasks, this long document benchmark features 5 repeated runs, so we have done so as well.

We describe the number of classes for each dataset in Table \ref{datasets} and the lengths of their documents in Table \ref{lengths}.

\section{Hyperparameters}

Table \ref{tab:hyperparameters} lists the hyperameters set for each experiment, primarily chosen by heuristics as follows. We have the size of the excerpt windows W, a stride S, the choice of selector (shared or feature tree) which has a corresponding classification layer. We mostly used the shared classifier since it generally performs better and requires less tuning ($n\_excerpts=classes$). For simplicity, we targeted a window size of 256 tokens with a stride at $\frac{1}{4}$ the window size for variation recalling strides in CNN's with shifting augmentation akin to translation augmentation in image processing \cite{shorten2019survey}. The so-called \textit{shared feature map} selector classifiers must have a number of excerpts directly proportional to the number of classes which can lead to dramatic increases in memory usage, mandating shorter excerpts for extreme classification problems as in the case of EURLEX with over 4000 classes. For CLEF23, we use a shared selector with a top-k value of 3 since it was trained on only 817 examples (news sources) where there are more biased instances (articles) to learn from each bag (source). The other exception for W is \textsc{PAN19-Birthyear}, which required the \emph{feature tree} selector because the \emph{shared selector} underperformed during the hyperparameter tuning runs. The variance in losses and reweighting is explained by the key performance metrics; for single-class datasets emphasizing macro-f1, we used class-based reweighting, but not for those emphasizing accuracy or micro-f1 (eg. longtext benchmark and CLEF23B). For multilabel datasets, we used ZLPR loss that balances multi-label loss with respect to class frequency\cite{su2022zlpr}. We trained models with model checkpointing for the key metric per dataset in tables (eg. accuracy, micro-f1, macro-f1). We used learning rate annealing. For the Gumbel-softmax, tau was set to 1 with the soft sampling technique and a decay rate of $\frac{1}{2}$ over 48 epochs.  We used RoBERTa large as the model to embed excerpts \cite{liu2019roberta}.  

\begin{table}[h]
\centering
\resizebox{\linewidth}{!}{\begin{tabular}{llllll}
\toprule
\small
\textbf{Dataset} & \textbf{Bag} & \textbf{Instance} & \textbf{Labels} & \textbf{ILabels} & \textbf{Type}\\
\midrule
Trigger Warning  & Stories & Passages & 32 & 8 &Multilabel \\  \hline
CLEF2023  & Sources & Articles & 3 & 3&Multiclass \\  \hline
Hyperpartisan & Articles & & 2 & -- & Binary \\ \hline
20News & Messages &  &20 & -- & Multiclass \\ \hline
Book Text & Summaries & & 227 & -- & Multilabel \\ \hline
Eurlex & Legal Acts & & 4271 & -- & Multilabel \\ \hline
PAN19-Gender & Tweets& & 3 & -- & Multiclass\\ \hline
PAN19-Fame & Tweets & & 3 & -- & Multiclass \\ \hline
PAN19-Occupation & Tweets & & 8 & -- & Multiclass\\ \hline
PAN19-Birthyear & Tweets & & 1 & -- & Regression\\ 
\bottomrule
\end{tabular}}
\caption{\label{datasets}
The datasets used in these experiments which include longer documents and collections of text. All include bag labels, while some also include instance labels.}
\end{table}

\begin{table}[h]
\centering
\small
\begin{tabular}{lrrrr}
\toprule
dataset & mean & median & 95th & max \\
\midrule
clef-23-3a & 98 & 115 & 152 & 387 \\
hyperpartisan & 749 & 544 & 1991 & 5560 \\
eurlex & 738 & 544 & 2000 & 6042 \\
book-text & 571 & 357 & 1741 & 13876 \\
20news & 355 & 210 & 930 & 28996 \\
clef-23-3b & 9173 & 7250 & 26993 & 77625 \\
trigger-warning & 3370 & 2964 & 7315 & 104470 \\
pan-2019 (all) & 77158 & 83727 & 134727 & 882343 \\
\bottomrule
\end{tabular}
\caption{\label{lengths}
The length of the datasets in RoBERTa tokens 
}
\end{table}
\section{Results}
We now describe the results of our experiments, starting with classification performance and memory and compute usage.  We follow with a visual analysis of how selector chosen instances and subsequent bag-level representations contrast with the other instances in a bag.  We conclude with an analysis of the the explanatory qualities of instances chosen by the selectors.

\subsection{Classification Results}

\begin{table*}[h]
\centering
\small
\resizebox{\linewidth}{!}{\begin{tabular}{lrrllrrlrrl}
\toprule
experiment & token\_max\_length & stride & selector & classifier\_type & n\_excerpts & num\_labels & loss & topk & batch\_size & reweight \\
\midrule
pan19-occupation & 256 & 64 & shared & shared & 8 & 8 & cross-entropy & 1 & 8 & True \\
pan23-triggers & 256 & 64 & shared & shared & 32 & 32 & ZLPR & 1 & 4 & False \\
pan19-birthyear & 64 & 16 & feature\_tree & feature\_tree & 8 & 1 & MSE & 1 & 32 & False \\
eurlex & 64 & 16 & shared & shared & 4271 & 4271 & ZLPR & 1 & 4 & False \\
hyperpartisan & 256 & 64 & shared & shared & 2 & 2 & cross-entropy & 1 & 8 & False \\
book-text & 256 & 64 & shared & shared & 227 & 227 & ZLPR & 1 & 8 & False \\
clef23b & 256 & 64 & shared & shared & 3 & 3 & cross-entropy & 3 & 8 & False \\
pan19-gender & 256 & 64 & shared & shared & 3 & 3 & cross-entropy & 1 & 24 & True \\
pan19-fame & 256 & 64 & shared & shared & 3 & 3 & cross-entropy & 1 & 8 & True \\
20news & 256 & 64 & shared & shared & 20 & 20 & cross-entropy & 1 & 8 & False \\
\bottomrule
\end{tabular}}
\caption{Hyperparameters for each experiment}
\label{tab:hyperparameters}
\end{table*}

First, we share results in MIL problems where a model trained on bag-level information achieves state-of-the-art results on that task but also generalizes to instance passage classification. We achieve state-of-the-art classification results for news source (CLEF 3B) which generalizes to accurately classify individual articles though it lags behind the state-of-the-art methods trained on the more plentiful, article-labeled data; see Table \ref{tab:clef}. In Table \ref{tab:trigger}, we share results for trigger-warning detection at the story-level (bag) as well as passage. We achieve state-of-the-art results at the bag-level and respectable accuracy at the passage-level. We present results on the PAN 2019 author profiling task where our solution creates new state-of-the-art results for gender and occupation but not age or fame in Table \ref{tab:long}. We report performance of our approach on long-document classification in Table \ref{tab:long}. We present an ablation study on 20News with 5 seeded runs for insight into hyperparameters in Table \ref{tab:hyperparameters}. For this dataset, we can see that window size of 256 performs better than 128, and the shared classifier performs best; feature tree does best with 6 excerpts, indicating potential overfitting with 4 more. 

\begin{figure}[h!]
    \centering
    \includegraphics[ scale=0.4]{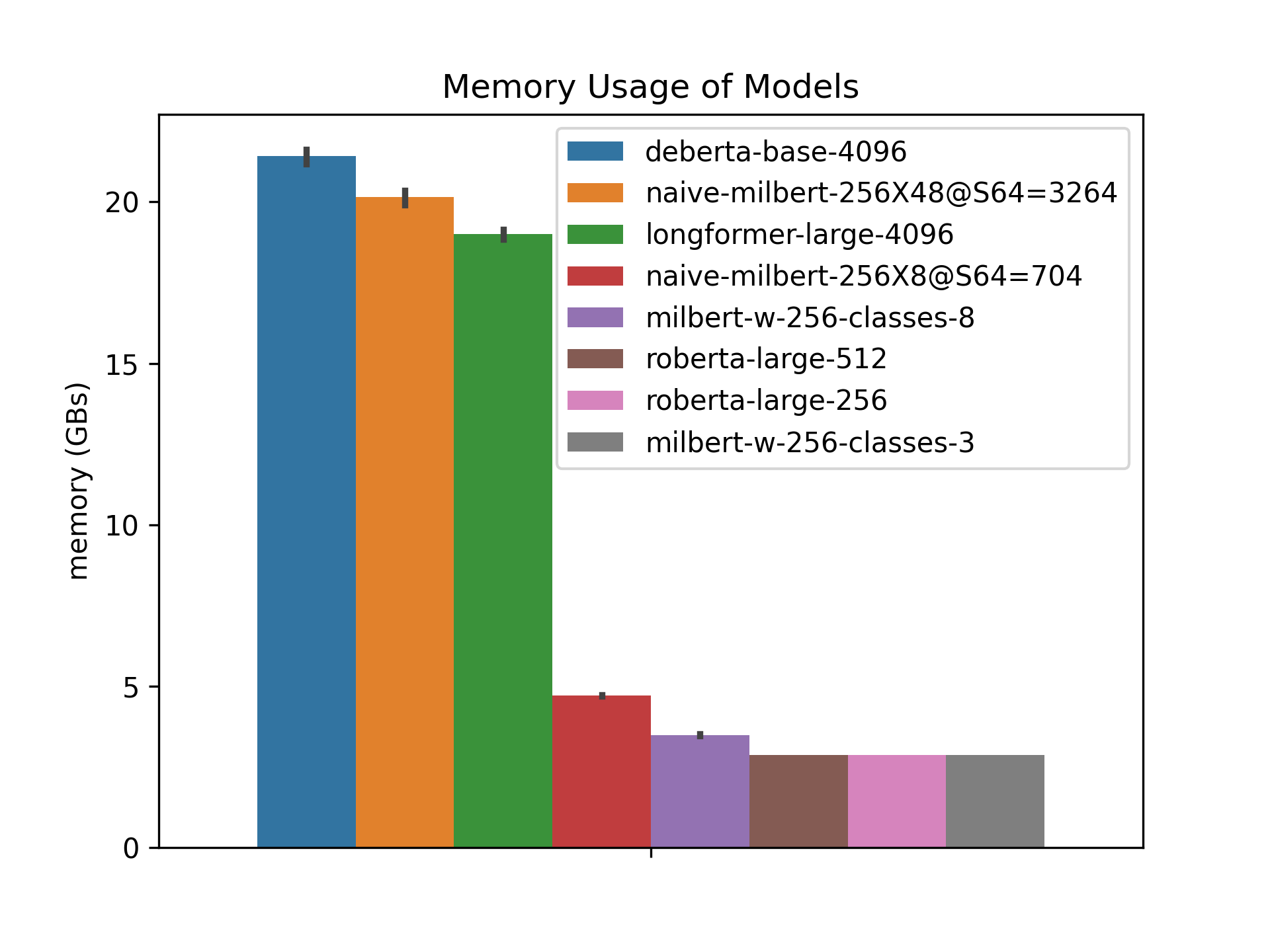}
    \includegraphics[ scale=0.4]{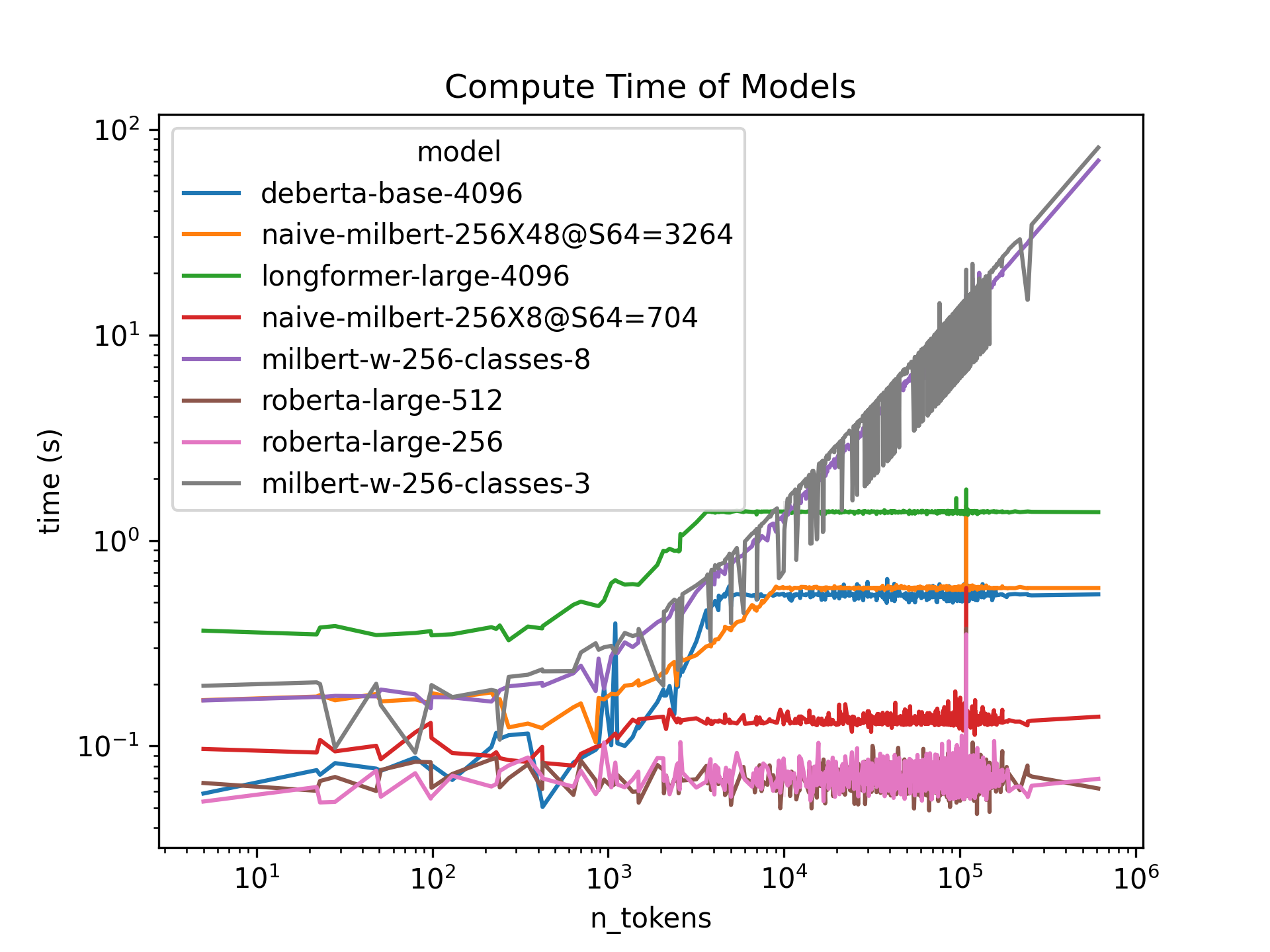}
    
    \caption{Memory usage of algorithms compared with truncation on the PAN 2019 dataset with batch size 1 and no optimizer. Truncation limits appear in all other approaches. Time to perform backpropagation calculated as the number of tokens varies.}
    \label{fig:memory}
\end{figure}

We present the memory and compute time profile of our approach compared with some popular approaches in Figure \ref{fig:memory}, including LongFormer \cite{beltagy2020longformer}. For each approach, we calculated the maximum resident VRAM during backpropagation on a single instance, clearing cached memory between each inference. Results conform with our description of $O(k)$ memory usage with $O(n+k)$ compute; in direct comparison, a naive max pooling MIL-BERT approach without the double pass pooling trick consumes 6 times the memory to consider only 48 contexts of 256 tokens compared to 8-class MIL-BERT. As for compute, we can see continual growth for longer documents for MIL-BERT while other approaches stop using more once they reach their truncation limit. Training on a single instance can approach 100 seconds for 1M tokens. Surprisingly, we observe that the fully trained 8-class MIL-BERT PAN19 occupation model uses less memory than the equivalent 8-excerpt, naive MIL-BERT, but single-passages may be selected multiple times by a selector, reducing memory requirements below the max in the second pass. Compared to other long-text approaches, MIL-BERT covers far more tokens than even long-context encoders such Longformer-large and DeBERTa base with much less memory usage. 

\subsection{Chosen Instances Visual Analysis}
\begin{figure*}[h!]
    \centering

    \includegraphics[ scale=0.4]{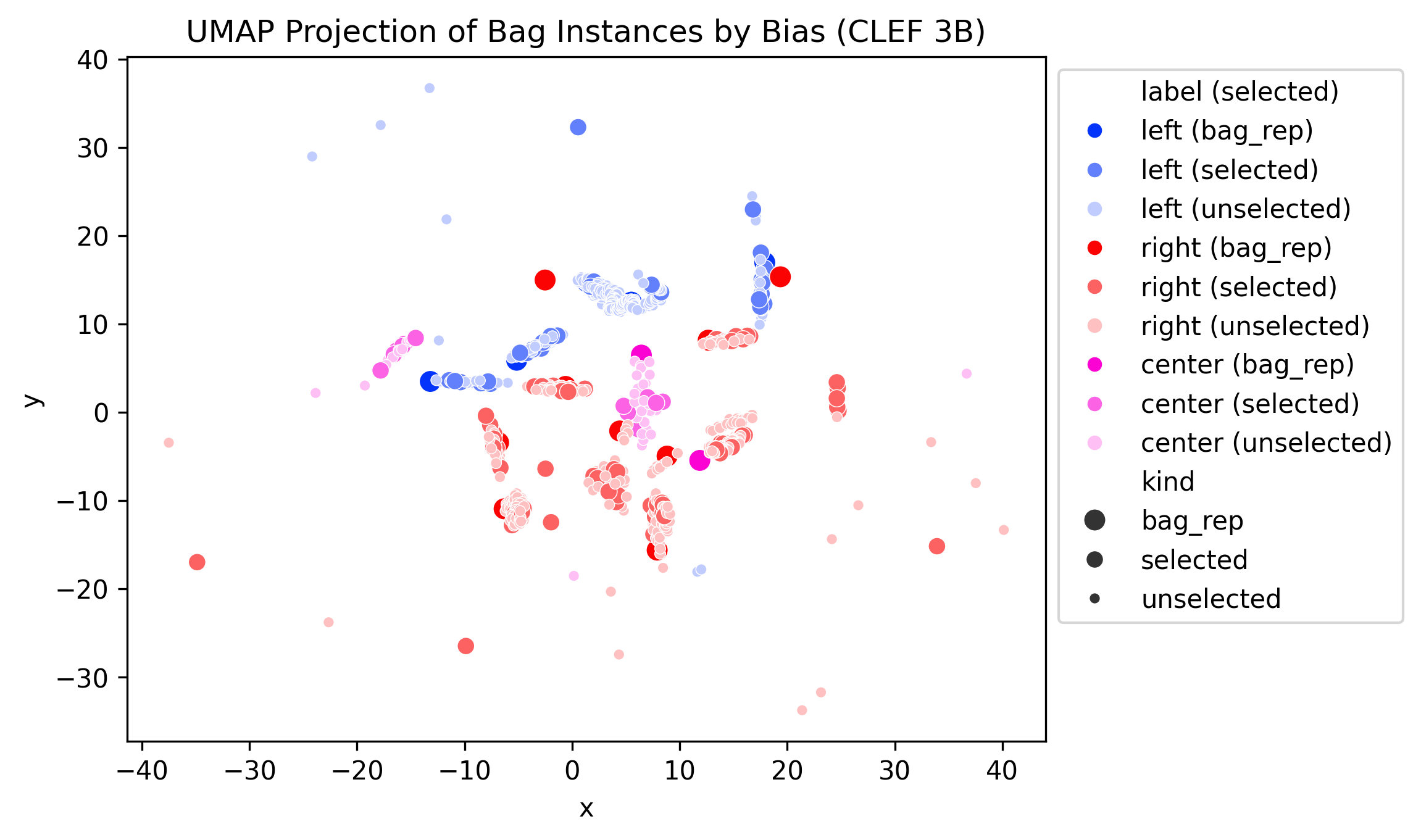}
    \includegraphics[ scale=0.4]{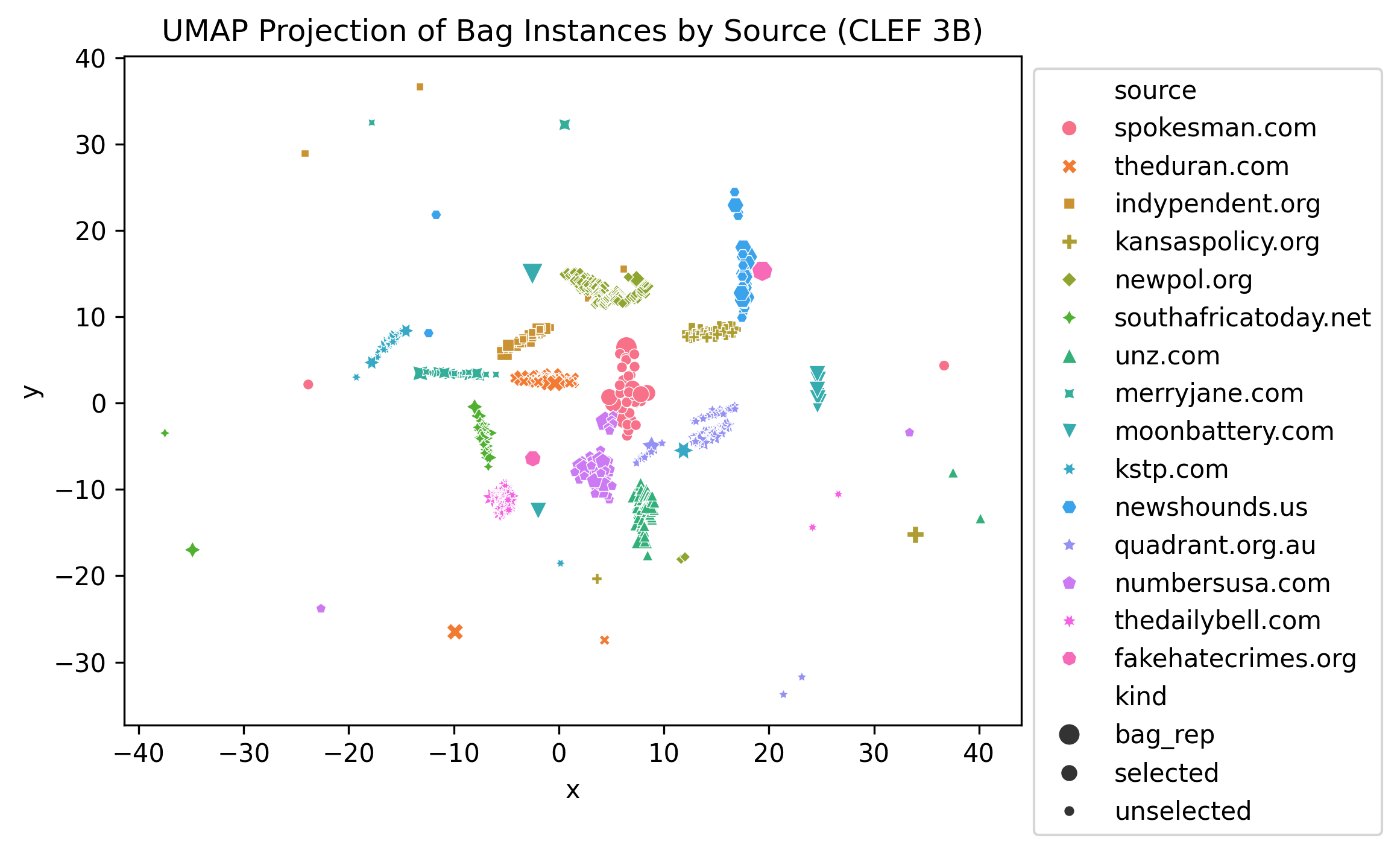}
    \caption{UMAP projection of collections of excerpts from various sources in the CLEF23 3B dataset with few neighbors to preserve local structure. The top 3 instances are selected per dimension and average to create the bag score. Selected instances are on the convex hull in classification boundary space of 3 dimensions.}
    \label{fig:bag-visualization}
\end{figure*}

We now present a visual analysis to contrast how selector chosen instances contrast with the other instances in a bag, starting with the shared classifier.
We present a UMAP projection~\cite{mcinnes2018umap} of the 3-dimensional logits for each instance (excerpts) from 15 sources scored with the shared classifier from the CLEF23 3B test set in Figure \ref{fig:bag-visualization} \cite{healy2024uniform}. 
To illustrate the local bag structure, we select 2 neighbors, a minimum distance of 0.05 and Euclidean distance as our metric. 
In addition to selected and unselected excerpts,  we present the bag representation which is the average of 3 points in each direction to create the logits for each bag. 
As would be expected, selected points line on the convex hull of each bag.  Given they contribute the largest feature score magnitudes, they would be the most descriptive for the classification boundary.
Instances of same source tend to be tightly grouped together--likely due to topical and ideological consistencies across documents as well as duplicate content between excerpts from striding. 

\subsection{Explainability}
By construction the feature selector picks instances with the largest magnitude along each feature dimension. 
In the classifiers used for this work, these instances are the ones that drive the final decision as they account for the largest classification score contribution across the feature dimensions.
Thus instances picked by the feature selector guaranteed to be the ones that would be most impactful on the classifier's decision.
Similarly, during learning they are also the ones that contribute most strongly to the weights update.

Given this, selector scores for excerpts provide qualitative insights about the decision function.
In addition, they can be analyzed quantitatively with correlation or conventional explainability techniques. 

\begin{figure*}[h!]
    \centering
    \includegraphics[ scale=0.45]{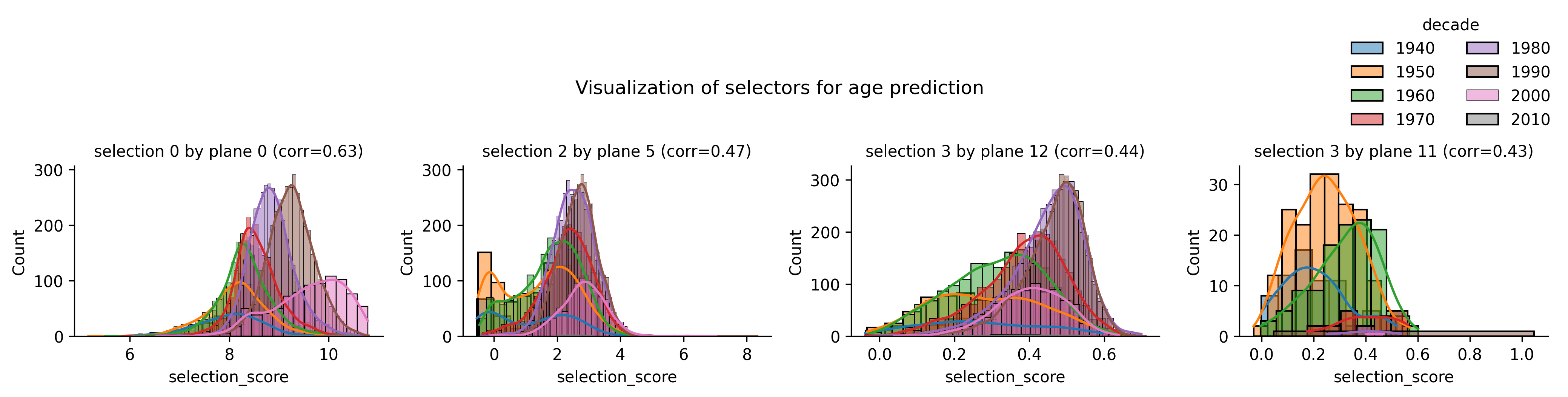}
    \caption{Visualization of selector scores from the age profiling feature tree model. Each histogram is a selector hyperplane and the X-axis is the selector score. Individual selectors specialize to features relevant in age profiling.}
    \label{fig:selector-visualization}
\end{figure*}

\begin{figure*}[h!]
    \centering

    \includegraphics[ scale=0.3]{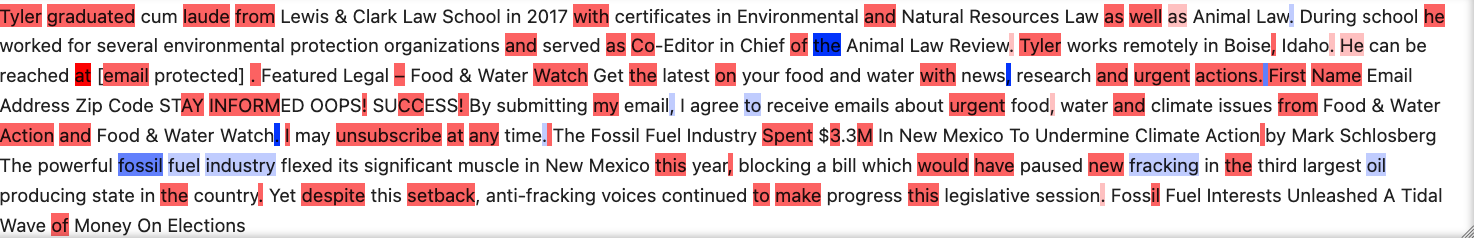}
    \caption{An excerpt classified as obvious left bias from a left-bias source with tokenwise attribution under layer integrated gradients. Blue indicates positive attribution toward the selector score, rather than the classification decision. White is near neutral and red contributes against the selector. This excerpt is heavily influenced by "fossil fuel industry."}
    \label{fig:explain-visualization}
\end{figure*}

First, we can analyze in aggregate by considering how their individual scores in each dimension (Figure \ref{fig:selector-visualization}) correlate with the classes of concern. We present selector dimensions of individual selector scores with separate histograms for each decade of birthyear. We calculate a moderate correlation between these selector scores and  birthyear (0.42-0.62). We notice that all excerpts are beyond plane 0 (positive) while plane 5 has scores above and below zero. This plane bifurcates authors loosely around 1970 and passes them on to planes 12 and 13, representing specialization like CNN feature maps. Second, individual excerpts and selector scores can be interrogated for explainability purposes with layer integrated gradients \cite{sundararajan2017axiomatic} as provided by Captum \cite{kokhlikyan2020captum}. In Figure \ref{fig:explain-visualization}, we present a visualization of the selected left-bias excerpt from a liberal news source by a shared selector. Gradients were calculated against the word embeddings to calculate how they impact the output of the selector, not the overall classification decision. Putting these methods together, it is possible to characterize how individual passages contribute to the final score and why they are selected.

\begin{table}[h]
\centering
\tiny
\begin{tabular}{lllll}
\toprule
Dataset        & \multicolumn{2}{l}{Bag (CLEF-3B)} & \multicolumn{2}{l}{Instance (CLEF-3A)} \\ \midrule  
               & MAE        & Macro-F1   & MAE         & Macro-F1       \\ \midrule 
Ours           &  \textbf{0.22}  &  0.87        &  0.69        & 0.39    \\ \midrule      
Frank23 3B       &  0.32     &  ---        &  ---      & --    \\ \midrule   
Frank23 3A       &  ---     &  ---        &  \textbf{0.27}        & --  \\    
\bottomrule
\end{tabular}
\caption{\label{tab:clef}
Results for the 2023 political bias dataset.
}
\end{table}

\begin{table}
\centering
\tiny
\begin{tabular}{lllll}
\toprule
Dataset        & \multicolumn{2}{l}{Bag} & \multicolumn{1}{l}{Instance} \\ \midrule  
               & Macro-F1   & Micro-F1   &  Average Acc    \\ \midrule 
\textbf{Ours}  &  \textbf{0.37}     &  \textbf{0.75}       &   0.709          \\ \midrule

\cite{sahin2023arc}  &  0.352     &  0.74    &   ---         \\ \midrule
5 fold multiclass &   ---      &   ---      &   0.699        \\ \midrule
Mixtral8x7B &   ---      &   ---      &   \textbf{0.747}       \\ 
\bottomrule
\end{tabular}
\caption{\label{tab:trigger}
The performance of classifiers trained on bag-level labels along with performance on instance-labeled datasets for PAN23 trigger warning dataset. Instance-level labels are majority vote decisions for 7 classes from the in-domain subset: violence, death, abduction, racism, homophobia, misogyny, ableism.
}
\end{table}

\begin{table}
\centering
\tiny
\resizebox{\linewidth}{!}{\begin{tabular}{lllll}

\toprule
Metric        & \multicolumn{2}{l}{Accuracy} & \multicolumn{2}{l}{Micro-F1} \\ \midrule  
Dataset               & Hyperpartisan  & 20News     & EURLEX   & Book Summary       \\ \midrule 
Ours             &   92.31 ($\pm5.2$)   &   85.25 ($\pm.7$)      &  69.21  ($\pm.6$)        &  58.16 ($\pm.5$)          \\ \midrule 
BERT            &    92.00        &   84.79    &  73.09     &   58.18          \\ \midrule
BERT+TextRank   &    91.15        &   84.99 &  72.87     &   58.94          \\ \midrule
BERT+Random     &    89.23        &   84.65  &  \textbf{73.22}     &   \textbf{59.36}   \\ \midrule
Longformer      &    \textbf{95.69}   & 83.39       &  54.53     &   56.53     \\ \midrule
CogLTX          &    94.77       &   84.63       &   70.13     &   58.27     \\ 
\bottomrule
\end{tabular}}
\caption{\label{tab:long}
The performance of classifiers for long document benchmark with its partitions with previous results\cite{park2022efficient}. Metrics averaged over 5 runs with different seeds. Our results include standard deviation, unlike previously reported benchmark results.}
\end{table}

\begin{table}[h]
\centering
\tiny
\label{tab:ablation}
\caption{Results from an ablation study on 20News dataset for 5 seeds}
\begin{tabular}{lrrrr}
\toprule
selector & n\_excerpts & token\_max\_length & accuracy & accuracy-std \\ 
\midrule
feature\_tree & 6 & 128 & 0.502 & 0.411 \\ \midrule
feature\_tree & 4 & 128 & 0.716 & 0.044 \\ \midrule
feature\_tree & 4 & 256 & 0.720 & 0.066 \\ \midrule
feature\_tree & 10 & 128 & 0.819 & 0.006 \\ \midrule
feature\_tree & 10 & 256 & 0.831 & 0.008 \\ \midrule
feature\_tree & 6 & 256 & 0.837 & 0.012 \\ \midrule
shared & 20 & 128 & 0.844 & 0.008 \\ \midrule
shared & 20 & 256 & 0.853 & 0.007 \\ 
\bottomrule
\end{tabular}
\end{table}

\begin{table}
\small
\centering
\begin{tabular}{llllll}
\toprule
        & \multicolumn{4}{l}{Macro-F1}  \\ \midrule  
Dataset:               & Gender  & Age     & Fame  & Occup  \\ \midrule 
Ours          &  \textbf{0.783}    & 0.571   &   0.43         & \textbf{0.549}     \\ \midrule
\citet{radivchev2019celebrity}      &  0.726 & \textbf{0.618}  & \textbf{0.551}    & 0.515    \\  \bottomrule
\end{tabular}
\caption{\label{tab:long}
Performance on the PAN 2019 celebrity profiling dataset. Age is scored as a modified classification task with a hit-window ranging from 2-6 years.}
\end{table}

\section{Discussion}

We have demonstrated that MIL-BERT is an efficient and explainable algorithm for long texts or \candr problems with a natural MIL structure, delivering competitive accuracy. We treat identification of trigger warnings, political bias and demographic characteristics as MIL problems for the first time to achieve new state-of-the-art results, and documenting the accuracy of instance classifiers trained only via bag labels where possible (trigger warnings and political bias detection). While setting a new state-of-the-art in these tasks, we present competitive results in the long document benchmark, where no single method is dominant. Finally, we demonstrate ways selector scores of excerpts can be interpreted individually and in aggregate.


We bring progress to three datasets that have proven unwieldy for conventional neural text classification techniques due to memory limits. In trigger warning detection, previous state-of-the-art methods either use a frozen RoBERTa model as a feature extractor \cite{sahin2023arc} or train a RoBERTa model on just the first and last 512 tokens \cite{su2023siamese}. In a previous version of this task, an XGBoost \cite{chen2016xgboost} baseline outperforms Longformer because truncation leads to missing trigger warnings \cite{wiegmann2023trigger}. Likewise, traditional feature engineering approaches led the top team to success in author profiling, with negative results for neural approaches from 3 out of six teams \cite{wiegmann2019overview}. Finally, we present improved results for source bias classification CLEF23 3B task \cite{da2023overview} where the previous best team \cite{azizov2023frank} used CatBoost \cite{prokhorenkova2018catboost}. We have shown progress in terms of task metrics but also in terms of finding a suitable neural model. 

Our success is limited on some of the tasks. In the long-document classification benchmark, our approach is best in only one dataset, but no existing approach is dominant. Based on author profiling results, we conclude that MIL-BERT is most suitable for MIL tasks and is otherwise merely competitive in long-text \candr problems. Though we have improved performance in predicting an author's gender or occupation, our approach still lags behind in predicting birthyear and fame, presumably because these tasks do not exhibit the signal locality for which MIL-BERT is designed.. Gender and occupation can be identified by excerpts, which borders on the task of self-disclosure detection \cite{umar2019detection}. Another limitation of this study is the exploration of hyperparameters. We have tuned hyperparameters for the birthyear dataset, but further tuning could still improve results. If we had normalized birthyear to match the hit-window function used in evaluation, we would likely improve performance by weighting discrepancies for younger authors more heavily. One reason for this under-exploration is that we cannot escape the fundamental limits to our algorithm. Computation time inevitably increases with document length and the number of excerpts increases the memory burden. 


We leave it to future work to improve selector and classifier layers. For example, we have experimented with auto-regressive selection with transformers in other work. At each timestep, the input is the previously selected instance and the output is the means to select the next instance. For now we recursively compute key-instances, but future work could detach gradients as the points are sampled for $O(n)$ compute, though not for all selection methods such as feature tree. As for classification layers, we have used transformers in other problems with more training data since this can overfit. There are obvious dangers of this algorithm for mining protected characteristics of individuals, but if offers new \candr applications across multiple documents such as entity typing \cite{del2015finet} and MIL-based, distant information extraction \cite{weber2020pedl}. The double-pass pooling technique can be applied to many problems besides \candr, especially those involving retrieval and matching tasks, which we also leave to future work. Towards facilitating replication and future work, we plan to release our library containing MIL-BERT.


\section{Related Work}

Taking inspiration from approaches in MIL and explainable models, we present a solution which offers scaling to long-document classification with explainability guarantees.

\subsection{Multiple-instance Learning}

MIL has been extensively applied in biology, computer vision, and natural language processing \cite{carbonneau2018multiple} on problems involving collections of labeled instances (molecular orientations, image subregions, or textual excerpts). There are approaches which emphasize accuracy on either the bag or the instance such as \cite{qu2024rethinking}. MIL with the accurate identification of key instances has long been applied to NLP problems such as event detection \cite{wang2016multiple}. There are true approaches to learning neural decision trees  (e.g. \cite{karthikeyan2021learning}) and they have been extended to MIL problems\cite{konstantinov2023multiple}, but they use all instances and self-attention to produce an aggregate score. PEDL applies a type of MIL to learn protein-protein interactions based on mentioned contexts, but the authors could sample at most 100 contexts, ostensibly due to memory limitations \cite{weber2020pedl}. Likewise, other authors train separate instance and bag models \cite{delucia2023multi}. Our approach enables the processing of arbitrarily large bags under fixed memory usage.

\subsection{Explainability}

There are many broad approaches to explainability in NLP: selector-predictor architectures (as here),  post-hoc input perturbation, analysis of attention, and methods which analyze gradients \cite{luo2024local}. Before the Gumbel-softmax became popular, the REINFORCE algorithm was used to overcome the inherent issues of calculating gradients over discrete selection to enable selector-predictor models \cite{lei-etal-2016-rationalizing}. The Gumbel-softmax trick has been used to select top-k phrases to use in calculating aspect-oriented sentiment analysis \cite{paulus2020gradient}. This technique has also been used to align textual phrases with image segments \cite{zheng2020hierarchical}. We view this work as continued exploration of the established explainability of MIL approaches to text classification such as event detection (eg. \cite{wang2016multiple}), but we also identify new performance guarantees in the selector-predictor paradigm with idempotent selectors. 

\subsection{Long-document classification}

In long document classification, most research has pursued increasing the length of transformers attention mechanism such as Longformer \cite{beltagy2020longformer} or the Infinite Memory transformer \cite{martins2022former}. By contrast, some work has pursued identifying key-excerpts. CogLTX resembles this work in that they find a memory efficient way to sample subsections of a document, but it uses two BERT models to do so: a judge which recalls selectors and a final reasoner model \cite{ding2020cogltx}. Without predefined excerpts, the judge determines tokens sequences, concatenates them and passes them into a second model. Furthermore, the unsupervised training method involves ablating subsections and calculating loss in order to rank relevant sections as compared to our geometric formulation with direct ties to the primary loss function. Similarly, the BERT+TextRank \cite{park2022efficient} baseline uses an unsupervised algorithm TextRank \cite{mihalcea-tarau-2004-textrank} to select relevant content to learn against. ToBERT uses state-of-the-art hierarchical document processing model with a transformer of BERT subsegments \cite{9003958}, but such approaches can be prohibitively expensive in terms of memory as in the trigger warning task \cite{sahin2023arc}. 

\section{Conclusion}

We have demonstrated an algorithm for learning to classify arbitrarily large collections of text which exchanges compute for memory savings with the double pass pooling trick and idempotent selectors. Along the way, we present new state-of-the-art results in trigger warning detection and political bias detection of sources, which can then be applied to classify constituent elements. We present limited improvements to authorship profiling and competitive results in long document classification. In the near-term, we would like to see these techniques applied to more MIL problems, even those outside of text. We have outlined some possible improvements to selectors and classifiers which we would like to apply to other problems with more data or other related problems such as retrieval.

\section{Limitations}
The instance labeled data for the MIL experiments regarding article bias and triggering passages are limited. For the sake of full comparison with other approaches, we have not identified whether the bags containing these instances are seen in training. Nevertheless, we believe this is reasonable since the bag-labeled training data does not label instances explicitly.


In-depth profiling of memory usage of algorithms was impeded by the tendency of pytorch to reserve blocks of memory between samples even with calls to release cached memory. We ran each model in its own process with a GPU and emptied cached CUDA memory: deleting items, running garbage collection, and resetting pytorch's statistics. However, memory reservation may make our numbers closer to the high-water mark of GPU memory usage. Also, it seems to be less accurate at the lower end of memory utilization. Memory and compute requirements do increase with more classes due to more selected excerpts, but we do not observe the same expected multiple when considering the 3-class MIL-BERT classifier where it uses marginally less memory than the 8-class classifier. We believe this is due to limited visibility into memory allocation and fragmentation.

\section{Ethics Statement}
The capability of the algorithm to learn against arbitrarily large data presented here creates opportunities for performing classification problems that are otherwise infeasible. By virtue of selecting passages, it is interpretable. However, improved performance on author profiling indicates it could be used to infer protected characteristics of individuals if presented with such training data.  At the same time, it presents capabilities identifying trigger warnings.

\section{Acknowledgments}
This material is based upon work supported by the Intelligence Advanced Research Projects
Activity (IARPA) under Contract \#2022-22072200004 and the National Science Foundation (NSF) under Contract \#49100422C0013. Any opinions, findings and
conclusions or recommendations expressed in this material are those of the author(s)
and do not necessarily reflect the views of the IARPA or NSF.

\bibliography{custom}
\bibliographystyle{acl_natbib}

\appendix

\section{Appendix}
\label{sec:appendix}

\subsection{Training details}

We trained our models on a variety of machines. For our longest text dataset, we used 6 A6000s for 3 days. For the evaluation of how much memory is required and training time of a single bag we used a single RTX 3090.

\end{document}